\documentclass[10pt,twocolumn]{article}
\pdfoutput=1

\usepackage{setspace}
\usepackage{geometry}
\usepackage{blindtext}
\usepackage{graphicx}
\usepackage[utf8]{inputenc}
\usepackage{mathtools}
\usepackage{amsmath, amsthm, amsfonts, amssymb,color}
\usepackage{tikz, subcaption, cite}
\usepackage{tabu}
\usepackage{}

\usepackage{epsfig} 
\usepackage{psfrag}
\usepackage{mathrsfs}
\usepackage{nomencl}   
\makenomenclature
\usepackage{multirow}
\usepackage{array,booktabs,arydshln,xcolor}
\usepackage{soul}

\usepackage{mathtools}
\usepackage{nccmath}
\usepackage{caption}
\usepackage{nccmath}
\usepackage{comment}
\graphicspath{{RL_Agent_Figs/}} 

\newlength{\figW}   
\usepackage{stackengine}

\usepackage[comma,numbers,square,sort&compress]{natbib}
\usepackage{amsmath}
\usepackage{amssymb}
\usepackage{float}
\usepackage{amsthm}
\usepackage{graphicx}
\usepackage{epstopdf}
\usepackage{color}
\usepackage{verbatim}
\usepackage{tikz,times}
\usepackage{xcolor}

\usepackage[ruled,vlined]{algorithm2e}
\SetKwComment{Comment}{$\triangleright$\ }{}

\makeatletter
\renewcommand*\env@matrix[1][*\c@MaxMatrixCols c]{%
  \hskip -\arraycolsep
  \let\@ifnextchar\new@ifnextchar
  \array{#1}}
\makeatother

\providecommand{\keywords}[1]
{
  \small	
  \textbf{\textit{Keywords---}} #1
}

\date{}

\begin{document}

\title{Soft Constrained Autonomous Vehicle Navigation using Gaussian Processes and Instance Segmentation}

\author{Bruno H. Groenner Barbosa,
        Neel P. Bhatt,
        Amir Khajepour, and
        Ehsan Hashemi
        
\thanks{This work has been submitted to the IEEE for possible publication. Copyright may be transferred without notice, after which this version may no longer be accessible. Bruno H. G. Barbosa is with the Department of Automatics, Federal University of Lavras, Brazil, and Department of Mechanical and Mechatronics Engineering, University of Waterloo, N2L3G1, ON, Canada; Neel P. Bhatt and Amir Khajepour are with the Department of Mechanical and Mechatronics Engineering, University of Waterloo, N2L3G1, ON, Canada; Ehsan Hashemi is with the Mechanical Engineering Department, University of Alberta, Edmonton, T6G1H9, AB, Canada (e-mails: brunohb@ufla.br; \{npbhatt, ehashemi, a.khajepour\}@uwaterloo.ca). This research was funded by the Coordination of Improvement of Higher Education Personnel – Brazil (CAPES),  the  Canada Foundation for Innovation, and Natural Sciences and Engineering Research Council of Canada.}}


\maketitle
\begin{abstract}
This paper presents a generic feature-based navigation framework for autonomous vehicles using a soft constrained Particle Filter. Selected map features, such as road and landmark locations, and vehicle states are used for designing soft constraints. After obtaining features of mapped landmarks in instance-based segmented images acquired from a monocular camera, vehicle-to-landmark distances are predicted using Gaussian Process Regression (GPR) models in a mixture of experts approach. Both mean and variance outputs of GPR models are used for implementing adaptive constraints. Experimental results confirm that the use of image segmentation features improves the vehicle-to-landmark distance prediction notably, and that the proposed soft constrained approach reliably localizes the vehicle even with reduced number of landmarks and noisy observations.
\end{abstract}


\section{Introduction} \label{sec:Intro}
Reliable and accurate navigation is a critical necessity for safe performance of mobile autonomous systems, such as in autonomous vehicles, service robots, and mobile robots for automated storage in warehouses. Leveraging a wide range of proprioceptive and exteroceptive sensory information, mobile autonomous systems enhance navigation and mapping algorithms for accurate motion planning, trajectory tracking, and robot stabilization algorithms \cite{levinson2010robust, li2019deep, li2020toward}. Integration of semantic knowledge of the scene, in which appearance-based features of the explored space are classified, improves reliability of navigation \cite{salas2013slam++, lateef2019survey}. In this regard, map-based methods are promising approaches for achieving high-accuracy localization \cite{kassas2020cellular, polemap2020}. Dense maps, such as Point Cloud Map (PCM) \cite{burgard2006}, or landmark maps \cite{trafficlight2019} are also widely used to perform navigation. The former requires a large amount of storage memory and may be not applicable for large scale environments, whereas the latter is more compact. Existing dense mapping approaches are computationally expensive for on-board computation in autonomous mobile robots and vehicles, limiting real-time capabilities. To resolve this, simplified multi-frame inference that decouples geometry and semantics are utilized \cite{dai2017bundlefusion, bloesch2018codeslam}.

In the map-based navigation approach, the selected features are expected to contain relevant prior knowledge about the static information the robot/vehicle may encounter in the environment. In the intelligent transportation setting, traffic signs \cite{trafficsign2015,sparsemap2019}, traffic lights \cite{trafficlight2019}, lanes locations \cite{roadmarkings2017,lin2020lane}, pole-like structures \cite{polelike2017}, curbs, and other road markings are normally used in this context \cite{survey2020ieee}. Besides, artificial landmarks such as QR-codes are also utilized to reduce ambiguity \cite{polelike2017}. For feature detection, semantic segmentation, and instance segmentation, Convolutional Neural Networks (CNN) have been widely utilized \cite{deephrr2020}.
Complex CNNs such as AlexNet \cite{alexnet}, VGGNet \cite{simonyan2015deep}, Inception \cite{szegedy2014going}, and Resnet \cite{he2015deep}, have been developed due to the increasing computational capability and the availability of large datasets. As opposed to object detection, in instance segmentation, besides detecting (objects') bounding boxes, a per-pixel segmented mask and classification is obtained that results in better time efficiency \cite{xie2020polarmask} and precise detection, especially for irregular objects \cite{qi2020pointins}. 
Indeed, existing approaches have reached real-time operation capability comparable to object detection performance \cite{Bolya_2020,tian2020conditional,wang2020solov2,lee2020centermask}. According to \cite{wang2020solov2}, the instance segmentation approaches can be grouped into three categories: \textit{i)} top-down, in which the object bounding box is detected (anchor-free or not) and then semantic segmentation is applied \cite{maskrcnn,Bolya_2020,xie2020polarmask,qi2020pointins}; \textit{ii)} bottom-up, where pixels are firstly labeled and then clustered \cite{deepwatershed2017,gao2019ssap}; and \textit{iii)} direct methods, where instance segmentation is performed directly without box detection or feature embedding \cite{wang2020solov2}.

The proposed and experimentally verified soft constrained navigation approach in this paper is considered a landmark-aided method (i.e., so-called a priori map-based localization \cite{survey2020ieee}) and is different from Simultaneous Localization and Mapping (SLAM) techniques \cite{orbslam22017}, Structure-from-Motion (SfM) methods \cite{sfm2016}, or GPS-IMU fusion systems \cite{caron2006}. An important feature of the proposed method is that it could be used together with other approaches to enhance state estimation and localization, such as an add-on approach discussed in \cite{bundle2019traffic,visionenhanced2020} to reduce possible drifts in SLAM. 
The main contributions of the paper are summarized as:
\begin{itemize}
    \item A generic feature-based navigation framework is developed and verified experimentally.
    \item Within this framework, a hybrid distance predictor is developed through a Gaussian Process Regression (GPR) model on an instance-based segmented landmarks; this is to increase the reliability of relative distance estimation for far landmarks as well as the close ones.
    \item A soft-constrained particle filter is designed for pose estimation and navigation using kinematic/geometric constraints and the hybrid GPR distance predictor. 
\end{itemize}
The remainder of the paper is organized as follows. Section \ref{Sec2-Seg&GPR} presents landmark segmentation and the hybrid GPR-based distance predictor. The constrained particle filter is developed in Section \ref{Sec3-CPF}. Experimental validation through an autonomous vehicle platform and corresponding discussions are provided in section \ref{Sec4-Results}. The conclusion is drawn at last.

\section{Segmentation and Distance Estimation} \label{Sec2-Seg&GPR}
\begin{figure*}[tb]
\centering
\includegraphics[width=1\linewidth]{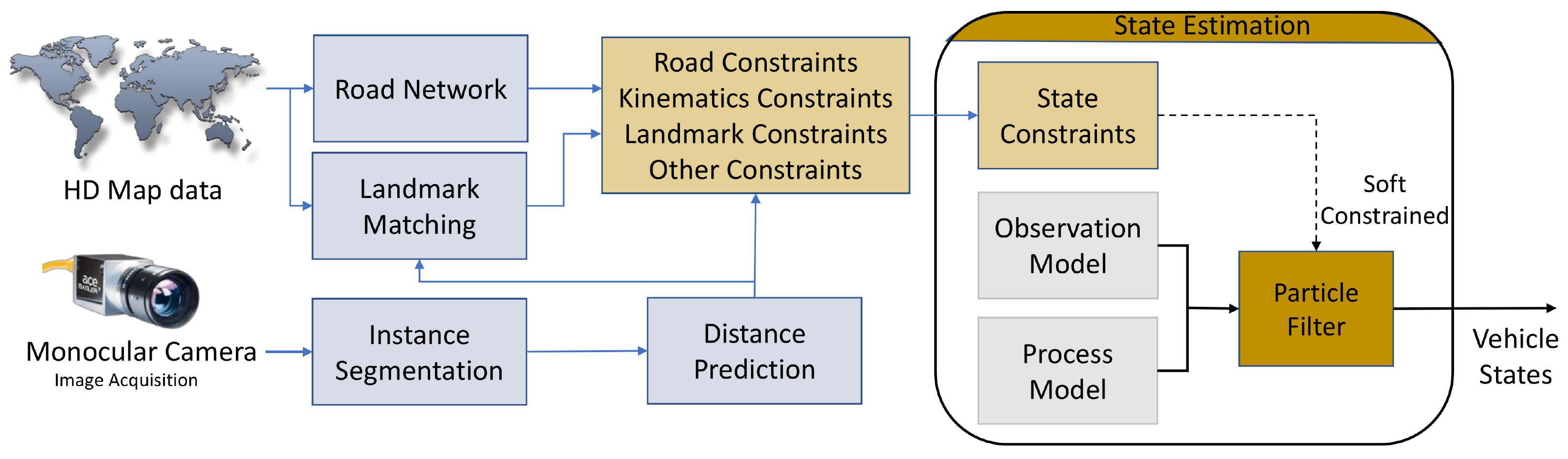}
\caption{General structure of the feature-based navigation algorithm.}
\label{fig:method}
\end{figure*}

An overview of the proposed localization framework is presented in Figure \ref{fig:method}. Firstly, global position of selected map features, such as landmarks and road central line position, are obtained. By means of a monocular camera, images are acquired and, based on an instance segmentation technique, known landmarks are detected and segmented (Sec. \ref{Sec4-GlobalPose}). According to features extracted from the detected and segmented landmarks, a Gaussian Process Regression model is implemented to predict the distance between the vehicle and each landmark (Sec. \ref{Sec2-Seg&GPR}). Considering these estimated relative distances, the detected landmarks extracted from images are compared to those from the available map and are properly associated (landmark matching process). With the predicted distances and the known road boundaries, constraints are developed to estimate the vehicle position based on a soft constrained particle filter algorithm (Sec. \ref{Sec3-CPF}).

The proposed localization framework relies on a predefined global positioning planar map of landmarks (Ground Control Points - GCP) and road shape (course) model, i.e., on a static HD map \cite{surveylocalization2018}. Thus, some relevant map features must be selected in order to generate the map before applying the proposed self-localization procedure. From this premise, a vision based semantic landmark search approach is employed here \cite{combininddeep2019lanmark}. 
Feature maps can be obtained using different approaches, such as by means of LiDAR and accurate GNSS systems or from public crowdsourced maps such as OpenStreetMap or even proprietary HD maps.
\subsection{Instance-Based Segmentation} \label{Sec-Instance}
In order to define landmark constraints for implementing the Soft Constrained Particle filter (Sec. \ref{Sec3-CPF}), previously selected map landmarks must be detected from RGB images and the relative distance between the vehicle and the detected landmarks have to be estimated. In this regard, for the sake of simplicity and considering that only planar HD map is available, vehicle, landmark and road elevations are neglected for relative distance prediction. Stereo-vision approaches \cite{stereo2006} and monocular depth estimation methods \cite{yin2019enforcing} are normally used to address scene analysis and calculate the relative positioning of objects therein \cite{geotagging2018}. However, considering that only one monocular camera is available, the proposed approach applies instance-based segmentation approach to obtain relevant features of known landmarks for predicting the relative distance between the vehicle and the detected and segmented landmark. 

\begin{figure*}[tb]
 \centering
  \includegraphics[width=1\textwidth]{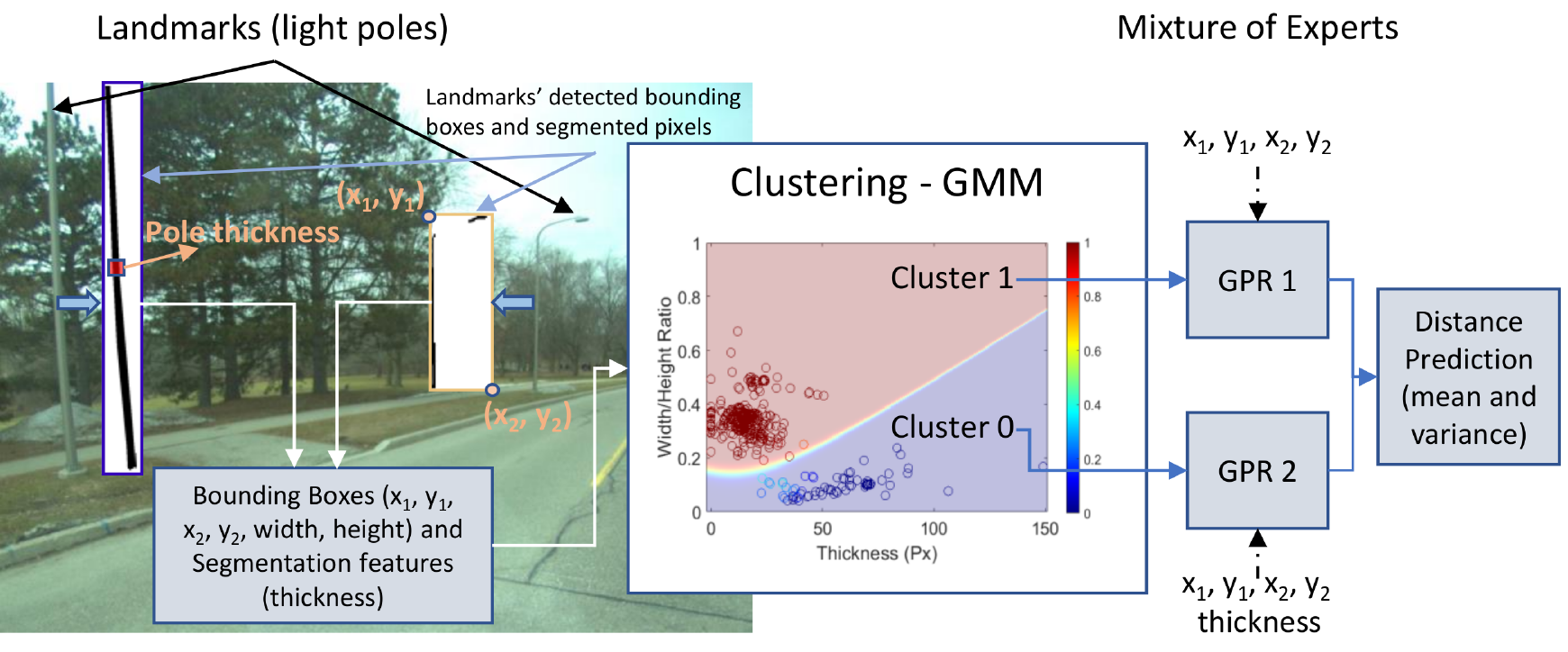}
  \caption{Relative distance prediction with GPR based on Landmarks (light poles) instance-based segmentation.}
 \label{fig:instance}
\end{figure*}

Figure \ref{fig:instance} presents an example of instance segmentation of landmarks (e.g. light poles). As can be noticed, besides providing the bounding boxes of the two detected landmarks, the algorithm also yields their associated pixel masks. We argue that the use of features extracted from the segmented landmark pixels together with bounding boxes features improve the object-to-camera distance prediction.

Some simple features such as bounding box height and width may be directly used for object-to-camera distance prediction. In some cases, these features would be enough since the detected object geometry is previously known and modification of the bounding box size is probably related to the change in the relative distance between the camera and the object. However, change in the bounding box size may be not related to object-to-camera distance due to camera distortion or even object occlusion. In this case, models based only on bounding boxes may fail. For instance, Figure \ref{fig:instance} presents two detected and segmented light poles (landmarks). It is worth noting that although the detected objects share the same geometry, their estimated bounding boxes as well as their segmented masks are of very different shapes. This occurs, in this example, not only because their distance to the camera are different but also due to the absence of the upper elbow (with light bulb) of the left light pole in the image. Besides, its bounding box is not coherent with the segmented pixels; it is wider than expected due to the camera distortion. Thus, other object features from the segmented mask can be included. For instance, the thickness of the left segmented pole (e.g. from number of pixels) could be obtained and also used as an input feature for the distance prediction model. This could improve the model accuracy and hence justifies the use of an instance segmentation algorithm. 

Despite all recent improvements on the algorithms for instance segmentation, the Mask R-CNN method \cite{maskrcnn} remains a versatile state-of-the-art technique \cite{tian2020conditional}. Briefly, Mask R-CNN is based on two-stage object detector, such as the Faster R-CNN \cite{ren2016faster}, to predict bounding-box for each instance and a compact fully convolutional network (FCN) for mask prediction, besides some operations on regions-of-interest (ROIs) of the network feature maps.

Thus, Mask R-CNN is used in this paper since our aim is to show the relevance of using instance segmentation to predict camera-to-object distance. Any other recent, more accurate and faster algorithm proposed in the literature could alternatively be promptly applied.

\subsection{Distance Prediction with GPR} \label{Sec-GPR}

After detecting and segmenting a known landmark, the next step of our approach is to predict the distance between the landmark (object) and the robot/vehicle in order to implement constraints for the localization algorithm, as shown in Figure \ref{fig:instance}. Considering real environments, the bounding boxes and the segmented masks of the detected objects are not perfect. For instance, partial object occlusion may occur or even the scene illumination may not be proper for correct object detection. 

In this way, some information about how reliable is the predicted distance should be used in the vehicle localization algorithm. Some papers have included uncertainties about object class identification for vehicle localization estimation, such as \cite{semantic2020,Ganti_2019}. However, they were applied to visual SLAM approaches and not for landmark-aided approaches such the one presented in this paper. Besides, the use of distance prediction uncertainty in vehicle localization algorithm were not found in the literature. 

Considering some landmarks of a map are well known, such as light poles or traffic signs, a regression model is implemented to predict the aforementioned distance with uncertainty. Each kind of landmark must have its own regression model since this model is designed using features of this specific landmark. One interesting approach for this task is the Gaussian Process Regression (GPR) model, a nonparametric probabilistic model based on Bayes theory \cite{rasmussen2006}.~An advantage of GPR is that the data define the input-output mapping complexity since no specific model structure is fitted \cite{barbosa20tire}.~Thus, GPR can handle complex relations between inputs and outputs with a relatively simple structure based on the mean and covariance functions \cite{SCHULZ20181}.~Providing the prediction variance value distinguishes GPR from other machine learning methods \cite{jin2015} and make it suitable for distance prediction in our proposed approach. 

In order to build a GPR model to predict the relative camera-to-object distance (desired output) considering some landmark features (input variables), assume that a data set $\mathcal{Z}$ is available such that $\mathcal{Z} \in  \mathbb{R}^{N \times (r+1)}$, where $\mathcal{Z} = [\mathbf{y} \ \ \mathbf{u_1} \ \ \mathbf{u_2} \ldots \mathbf{u_r}]$, $\mathbf{y}$ represents the known euclidean distance between the landmark and the camera (for instance in UTM coordinates), $\mathbf{u}$ represents the extracted features, $N$ is the number of samples (detected landmarks available in acquired frames) and $r$ is the number of features used as model's inputs. 

Consider that a function $f$ is distributed as a Gaussian process, and that $\mathbf{y} = f(\mathbf{u}) + \mathbf{\epsilon}$, with $\epsilon \sim \mathcal{N}(0, \sigma_\epsilon^2$) as the i.i.d noise term. The observed target prior distribution can be described as $\mathbf{y} \sim \mathcal{N} (\mathbf{0},\mathbf{K}(\mathbf{X},\mathbf{X}) + \sigma_\epsilon \mathbf{I})$ where $\mathbf{K}(\mathbf{X},\mathbf{X}) \in \mathbb{R}^{N \times N}$ is the covariance matrix between all observed data and $\mathbf{I}$ is the identity matrix \cite{barbosa20tire,rasmussen2006}.

To make a prediction $f_*$ for a new input vector $\mathbf{x_*}$, the joint distribution of the observed target values and $f_*$ follow a joint (multivariate) Gaussian distribution given by
\begin{align}
\begin{bmatrix}
\mathbf{y}\\
f_*
\end{bmatrix} \sim & \mathcal{N} \begin{pmatrix}  \mathbf{0}, \begin{bmatrix}
\mathbf{K}(\mathbf{X},\mathbf{X}) + \sigma_\epsilon \mathbf{I} & \mathbf{K}(\mathbf{X},\mathbf{x_*})\\
\mathbf{K}(\mathbf{X},\mathbf{x_*})' & \mathbf{K}(\mathbf{x_*},\mathbf{x_*})
\end{bmatrix} \end{pmatrix}. \nonumber
\end{align}

The mean and variance values of the posterior distribution $ P(f_*|\mathbf{X},\mathbf{y},\mathbf{x_*})$ are expressed as, respectively,  
\begin{align}
    {\mu}_* & = \mathbf{K}(\mathbf{x_*},\mathbf{X})[\mathbf{K}(\mathbf{X},\mathbf{X}) + \sigma_\epsilon \mathbf{I} ]^{-1} \mathbf{y}, \label{eq:gpr1}\\
    {\sigma}_{f_*}^2 & =  - \mathbf{K}(\mathbf{x_*},\mathbf{X}) [\mathbf{K}(\mathbf{X},\mathbf{X}) + \sigma_\epsilon \mathbf{I} ]^{-1} \mathbf{K}(\mathbf{x_*},\mathbf{X})  \nonumber\\
    & \ \ \ + \mathbf{K}(\mathbf{x_*},\mathbf{x_*}).\label{eq:gpr2}
\end{align}

To improve the GPR model accuracy (same object can have very distinct features according to the scene position as shown in Figure \ref{fig:instance}) and to reduce its computational complexity for real-time application (due to matrix inverse computation) a Mixture of Experts approach is employed \cite{liu2020gprscale}. In this case, the whole input space is divided by a gating network into $N$ smaller sub-spaces within which a simpler GPR model (expert) is used. The gating network can be seen as a clustering algorithm and different methods may be used, such as the Gaussian Mixture Model (GMM). The output of the ME approach could be a weighed combination of the $N$ experts outputs or even the output of the most appropriate one as implemented in this work. 

Note that the coordinate frames involved in the labelling of the training data as well as for real-time implementation includes: GPS, camera, and image frame. Additionally, knowing the transformation between these frames allows for local to global frame conversion and vice-versa.

After predicting the object-to-camera distance and considering that the camera coordinates origin is the lens optical center, this predicted distance should be converted to object-to-vehicle body coordinate. Considering the vehicle body is rigid and the camera is installed in a fixed and known position, the distance between the object and the vehicle body coordinates origin can be obtained regarding the bearing angle between the camera and the landmark is also known. To obtain the referred angle, it is necessary to convert image frame positions (pixel coordinates) to the camera coordinates, this can be done by intrinsic parameters calibration using chessboard, as suggested in \cite{trafficlight2019}.

\section{Vision-Map Localization} \label{Sec3-CPF}
After obtaining the landmark-to-vehicle relative distance, the landmark detected in an image must be associated to a landmark present in the available map (database), this is known as the association or landmark matching problem \cite{polemap2020}.  An approach to landmark matching is to apply a nearest neighbor search around the estimation of the vehicle’s pose in preceding frames \cite{trafficsign2015}. If more than one match is found, the predicted landmark-to-vehicle relative distance can be used to find the proper correspondence \cite{polemap2020}. However, this method may be unreliable when there are many landmarks in the map (dense map), or when the vehicle has a large displacement. The matching problem is widely discussed in the literature and it is not the focus of this work.
\subsection{Soft Constrained Particle Filter Design}\label{sec:scpf}
The vehicle motion is usually represented in a discrete form by a dynamic state model (or process model) describing how the vehicle states evolve over time and an observation model (or measurement model) that describes the relation between the states and the available observations, such that

\begin{align}    
\mathbf{x}_{k+1} &= f_k(\mathbf{x}_k) + \mathbf{w}_k, \nonumber \\  
\mathbf{z}_{k} &= h_k(\mathbf{x}_k) + \mathbf{v}_k,  \label{eq:ss}   
\end{align}   

\noindent where $f:  \mathbb{R}^n  \rightarrow \mathbb{R}^n$ is the process model and $h:  \mathbb{R}^n \rightarrow \mathbb{R}^m$ is the observation model, $ \mathbf{x}_k \in \mathbb{R}^n$ is the state vector, $ \mathbf{z}_{k} \in \mathbb{R}^m$ are  measured outputs, $ \mathbf{w}_{k} \in \mathbb{R}^n$ and $ \mathbf{v}_k \in \mathbb{R}^{m}$ are zero-mean uncorrelated process and measurement noises, respectively. Uncertainty models are included in order to deal with unmodeled dynamics, perturbations or measurement errors \cite{merlinge2019boxPF}. In this context, Bayesian filters are of paramount importance, \textit{e. g.} Kalman Filter (KF) and Particle Filter (PF), and applied to estimate the posterior distribution over the current vehicle state $\mathbf{x}_k$ denoted as $p(\mathbf{x}_k|\mathbf{z}_{1:k})$.

In this paper, Particle Filter is applied to vehicle position estimation due its capability of dealing with nonlinear and non-Gaussian dynamic systems \cite{arulampalam2002PF}. Aiming at improving the localization estimation, additional information about the system besides the vehicle fundamental dynamic are incorporated as states constraints in the state estimation problem, restricting the estimation to a certain feasible region. The use of state constraints may improve the accuracy of the estimation and reduce its uncertainty degree \cite{liu2019cPF}.

This information could be directly extracted from the available HD map, such as the road network (\textit{Road Constraints}). It could be also related to known kinematics constraints of the object, physical laws, speed constraints, among others (\textit{Kinematic and Other Constraints}) \cite{SHAO2010143}. Particularly, we are interested in additional external knowledge obtained from the vehicle perception system, \textit{i. e.} information extracted from images acquired by the monocular camera. More precisely, the objective is to implement state constraints by means of the predicted relative distance between the detected objects (matched landmarks) and the vehicle (\textit{Landmark Constraints}). 

State constraints are normally defined as a set of linear or nonlinear inequalities which can be expressed as,

\begin{equation}
    C_k(\mathbf{x}_k) \leq 0, \label{eq:hc}
\end{equation}

\noindent where  $C:  \mathbb{R}^n  \rightarrow \mathbb{R}^{n_c}$ are hard constraint functions at time $k$, and the inequality sign holds for all elements.

Considering that there are also uncertainties related to the constraints, such as in the landmarks global position or in the predicted relative distance, designing soft constraints instead of hard constraints is advisable, as discussed in \cite{liu2019cPF}. In this case, and for convenience, Eq. \ref{eq:hc} can be replaced by \cite{liu2019cPF}:

\begin{equation}
    C_k(\mathbf{x}_k) - \Gamma_k \leq 0, \label{eq:sc}
\end{equation}

\noindent where $\Gamma_k \in \mathbb{R}^{n_c}$ is an unknown vector of nonnegative
random variables, $\gamma_{i,k}$ with $i = 1 \ldots n_c$, that follows a pdf $p_\gamma(\Gamma_k)$, which makes Eq. \ref{eq:sc} nondeterministic. Considering the random variables $\gamma_{i,k}$ are independent of each other, $p_{\gamma_k}(\Gamma_k) = \prod_{i=1}^{n_c} p_{\gamma_{i,k}} (\gamma_{i,k})$, and the pdf  $p_{\gamma_{i,k}} (\gamma_{i,k})$ must be defined prior to state estimation. Two distributions are considered in this work, although others could be also selected. The first one is the exponential distribution with mean $\mu$ defined as,

\begin{equation}
\label{eq:exp}
    p (\gamma) = \begin{cases}
    \mu^{-1}\  e^{-\gamma/\mu}       &  \textrm{if} \ \ \ \gamma \geq 0 \\
    0  & \textrm{if}\ \ \  \gamma < 0
  \end{cases}
\end{equation}

\noindent that may be used for road and speed constraints, as suggested by \citep{liu2019cPF}. The second one is the zero-mean and truncated Gaussian distribution with variance $\sigma^2$, given by

\begin{equation}
    p (\gamma) = \begin{cases}
    2 \left ( \sqrt{2\pi}\  \sigma \right )^{-1}  e^{-\gamma^2/2\sigma^2}       &  \textrm{if} \ \ \ \gamma \geq 0 \\
    0  & \textrm{if} \ \ \ \gamma < 0
  \end{cases}
  \label{eq:gauss}
\end{equation}

\noindent that can be promptly applied to the landmark constraints since the proposed GPR model already provides the variance of the predicted relative distance that could replace $\sigma^2$ in the above equation. It is worth mentioning that this approach makes the soft constraint adaptive to the relative distance prediction uncertainty, relaxing the constraint when the prediction model is not reliable or tightening it in the opposite scenario.

\begin{algorithm}[t]
 \footnotesize{
 \caption{{Soft Constrained Navigation}}
 \label{alg1:distance_pred}
 \KwIn{\begin{itemize}
     \item Data set: $E$ consisting of images 
     \item Road Boundaries (polynomials)
     \item Kinematics Constraints
     \item GPS co-ordinates of features for Landmark Constraints
     \item State Constraints
     \end{itemize}}
 \For{img $\in E$}{
  \Comment{Distance Prediction:}
  (bbox,mask) $\gets instanceSeg$(img)\;
  $Cluster_i \gets \textrm{GMM}$(bbox,mask)\;
  ($\mu$,$\sigma$) $\gets \textrm{GPR}_i$(bbox,mask) from Eq. (1) and (2)\;
  \Comment{State Estimation - scPF:}
  Compute $p(\gamma)$ from Eq. (6) for road and kinematic (e.g. speed) constraints\;
  Compute $p(\gamma)$ from Eq. (7) using $\mu$ and $\sigma$ for landmarks based constraints - i.e. distance predictions\;
  $p_\gamma(\Gamma_k) \gets$ Stack the obtained constraints\;
  $\textbf{x}_k \gets$ Estimate the state using a Bayesian filter (e.g. PF) with the stacked constraints: $C_k(\textbf{x}_k)-\Gamma_k \leq 0$ and the process model in Eq. (3)\;
 }
 \KwOut{$\textbf{x}_k$: Vehicle position and velocities} 
 }
\end{algorithm}

Thus, assuming that $M$ matched landmarks with known global localization are detected in image $I_k$, acquired at discrete time $k$, and their distance to the vehicle is already predicted by the GPR model with mean and variance, and assuming that other external knowledge may be also available, such as vehicle maximum speed and road network, the problem is to estimate the vehicle localization at time $k$ given the HD map, the system state space-model and the designed soft constraints. In order to solve this state estimation problem, the Soft Constrained Particle Filter (scPF) proposed in \cite{liu2019cPF} is implemented. 

A remarkable feature of the proposed approach is that it can be used together with other localization methods, such as SLAM. Moreover, if there is not landmark detected in the current camera image, the state estimation proceeds normally as an unconstrained approach. The proposed approach is summarized in Alg.\,\ref{alg1:distance_pred}.

\begin{figure*}[t]
\centering
\begin{tabular}{c}
\includegraphics[width=1\linewidth]{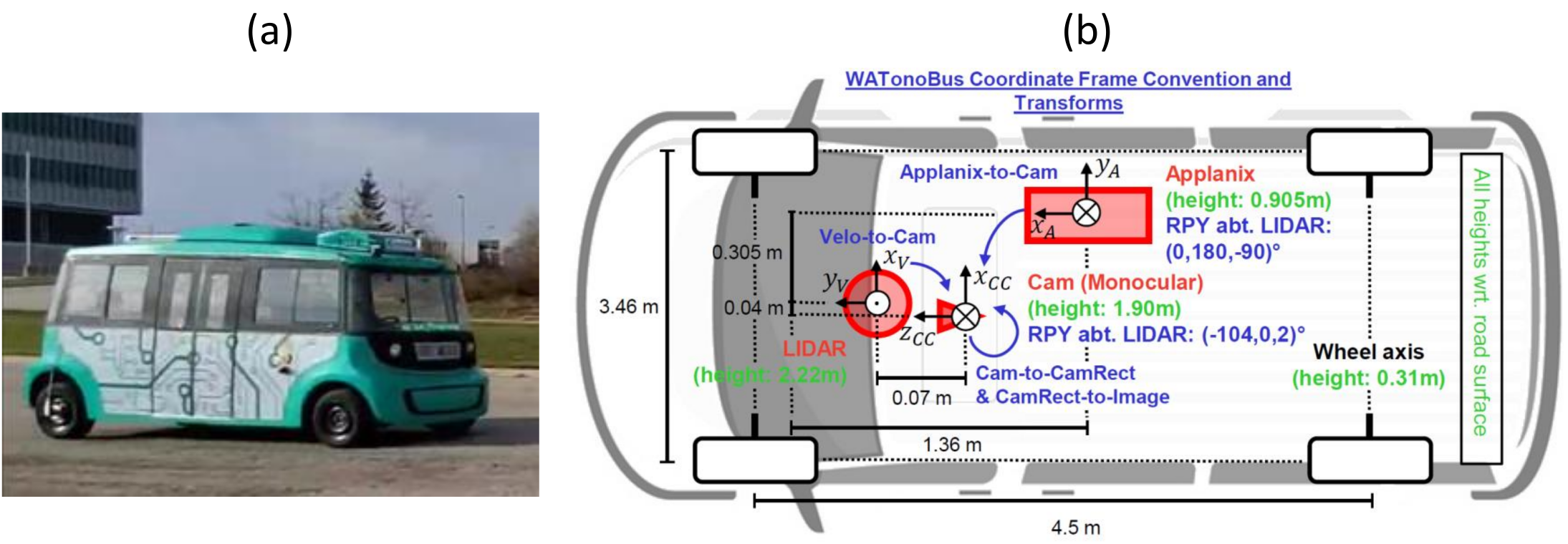} 
\end{tabular}
\caption{WATonoBus: (a) photo of the bus, and (b) dimensions (in meters) and position of the sensors (red) relative to the vehicle body -- heights above ground (green) are measured with respect to the road surface, transformations between sensors are shown in blue. Note that the LiDAR frame is used as a reference for measurements, but is not required by the algorithm. However, it might be required if the approach in Figure 10 is pursued for further enhancement.}
\label{fig:shuttle}
\end{figure*}

\section{Experiments and Discussions} \label{Sec4-Results}

The proposed framework was evaluated using a data set acquired from WATonoBus, the University of Waterloo autonomous shuttle bus (Figure \ref{fig:shuttle} (a)). WATonoBus is an electric vehicle with a capacity of 7 passengers. Its main dimensions, the sensors mounting positions, reference frames, and transforms are shown in Figure \ref{fig:shuttle} (b).  The measurements from the monocular camera and the GPS+IMU form the necessary sensory inputs. In our experiments, we used a 3.2MP Basler camera and Applanix POS LVX to obtain these measurements respectively. 

The shuttle bus was driven on part of the university Ring Road (a route of 1.2\,km concluded in approximately 5 minutes as shown in Fig. \ref{fig:RingRoad_Constraints} (a)), and a data set composed of 4000 samples was built. Ground truth of the shuttle bus position was acquired from Applanix POS LVX to calculate the algorithm errors and also to yield position measurements (observation data) to emulate rough GPS position data corrupted by a white Gaussian noise with variance $\sigma_v^2$. Images were acquired from the monocular camera, and the overall sampling frequency was 13\,Hz during the experiment.





\subsection{Map Features} \label{Sec4-GlobalPose}

\subsubsection{Ring Road and Poles Localization}

\begin{figure*}[tb]
 \centering
  \includegraphics[width=1\textwidth]{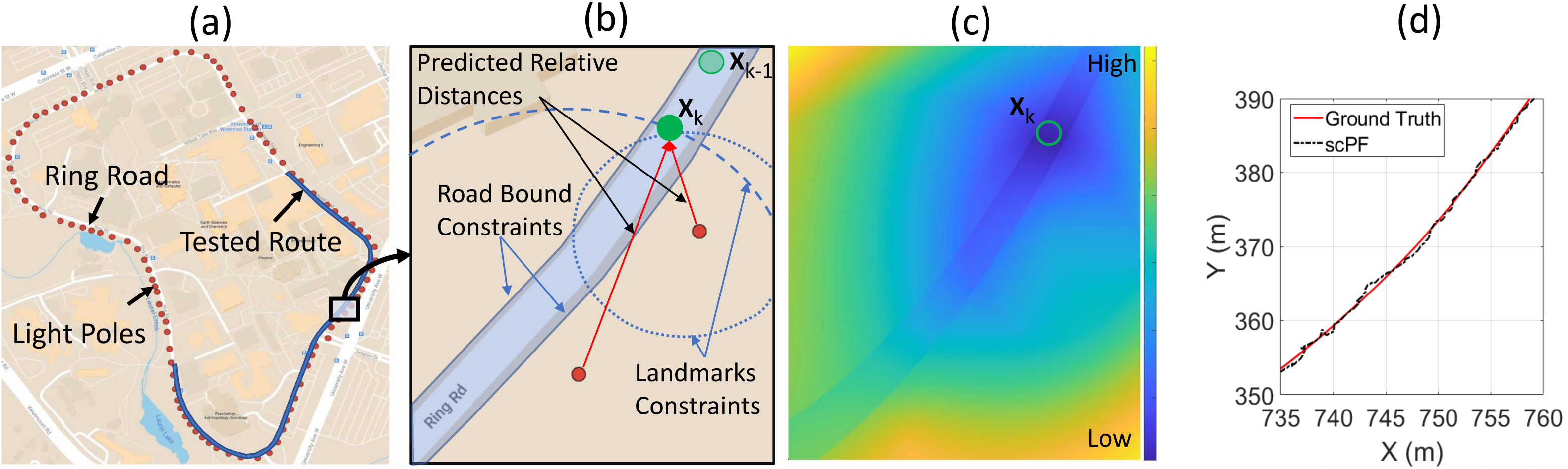}
  \caption{Ring Road and light poles localization (a), details of constraints designing (b), constraints' surface response (c), and some prediction of the proposed scPF localization approach (d).}
 \label{fig:RingRoad_Constraints}
\end{figure*}

It is important to have an accurate position of the poles locations for GPR models training purposes and also for the algorithm operation. However, in order to make our approach more practical, we obtained the 99 Ring Road poles locations directly from Google Maps although we know this may lead to errors in the range of 1-2 meters. Besides, the Ring Road center line was obtained directly from Applanix POS LVX measurements. The Ring Road and mapped Light Poles are shown in Figure \ref{fig:RingRoad_Constraints} (a). 




\subsubsection{Poles Detection and Segmentation}

In this work, Mask R-CNN instance-based segmentation algorithm was implemented to detect and segment Ring Road light poles in the monocular camera frames. To this end, 250 images out of the collected 4000 images were labeled (pole's bounding box and segmentation) using Labelme annotation tool, from which 200 images (or 314 poles instances) were used for model training and the remaining 50 (or 72 poles instances) for model validation. We fine-tuned a Mask-RCNN with Resnet 50 and Feature Pyramid Network (FPN) backbone (ResNet-50-FPN) pretrained on COCO (Common Object in COntext) dataset \cite{maskrcnn}. This procedure was realized on Detectron2, an open source PyTorch based modular computer vision model library developed by the Facebook AI Team. 
After fine tuning the model over 10000 epochs, no false positive was detected by the model on both training and validation data, the threshold used for detection was 0.9.  Besides, 16 out of 386 (6\,\%) poles were not detected (false negatives), 10 of training data and 6 of the validation one. These not detected poles were far from the vehicle, making the detection task more difficult. It is important to emphasize that precision is more important than recall in our approach since a false positive would add an erroneous constraint. Figure \ref{fig:instance} presents two light poles detected and segmented by the trained Mask-RCNN in a sample image. As a role of thumb, we observed that bounding boxes were well predicted in all samples, but closer poles (those with vanished light bulbs) were much better segmented than the others. This is expected since closer poles are much well defined in a pixel-level, which may be useful to predict the distance between these poles an the vehicle, as discussed next.

\subsubsection{Poles Distance Prediction}

In order to predict vehicle-to-landmark distance, a mixture of experts approach composed of two GPR models was developed. The implementation of this approach has two motivations: \textit{i.} different models should be used since detected landmarks shapes may vary (for instance, bounding boxes' height and width ratios or the availability of segmented pixels could be very distinct as shown in Fig. \ref{fig:instance}); \textit{ii.} GPR models may be very computational demanding if large data sets are used to train, although this is not an issue in the present work. 

For developing the mixture approach, we implemented a Gaussian Mixture Model (GMM) clustering algorithm to find two clusters, one to represent the closest poles (without lamp bulbs) and another related to poles that are far from the vehicle (normally with lamp bulbs). We defined two features as inputs for the clustering technique: ratio between bounding box's width and height and pole thickness (see Fig. \ref{fig:instance}). Thus, the GMM was trained using the training output of Mask-RCNN, during 1000 iterations, considering diagonal co-variance. 

Although we have implemented an unsupervised method, we also labeled the poles in order to validate the clusters found by the GMM. In this case, poles with light bulbs were defined as 1s, poles where the upper parts are completed vanished were designed as 0s, and poles which the upper part does not contain the lamp bulb \textit{and} is not completely vanished, we attributed numbers between 0 and 1, accordingly. Figure \ref{fig:instance} shows poles assigned values used to verify the clusters separation surface. As can be observed, the GMM separation surface  defined well the two kinds of detected poles (poles are assigned to the cluster with highest posterior probability), and those poles we labeled with values different from 0 or 1 are normally found close to the separation curve (the color bar represents our labeled values and poles which assigned values are greater than 0.5 should be classified as 1 and \textit{vice-versa}).

After clustering the poles, two GPR models were trained, one for each cluster. Both were trained using a holdout cross-validation (CV) procedure (100 repetitions) where 80\% of data were used for training and the remaining 20\% for validation (training data was the same as of Mask-RCNN). The selected kernel function was from class \textit{Matérn} with parameter $v=3/2$ and with automatic relevance determination (ARD) \cite{rasmussen2006}:
\vspace{0.08cm}

\begin{equation}
   \kappa_{v = 3/2}(\tau) = \sigma_f^2\left( 1 + \sqrt{3}\frac{\tau}{l}\right ) \rm{exp}\left(- \sqrt{3}\frac{\tau}{l}\right ),
\end{equation}
\vspace{0.0cm}
\noindent where hyperparameters $\mathbf{\theta} = \{ \sigma_f^2, l\}$ are positive, $\sigma_f^2$ is the output scale and $l$ is the input scale. The hyperparameters are obtained through a gradient-ascent based optimization tool, by maximizing the log-likelihood of the training data \cite{rasmussen2006}.

\begin{figure}[ht]
\centering
\begin{tabular}{c}
(a) \\ \includegraphics[width=0.95\linewidth]{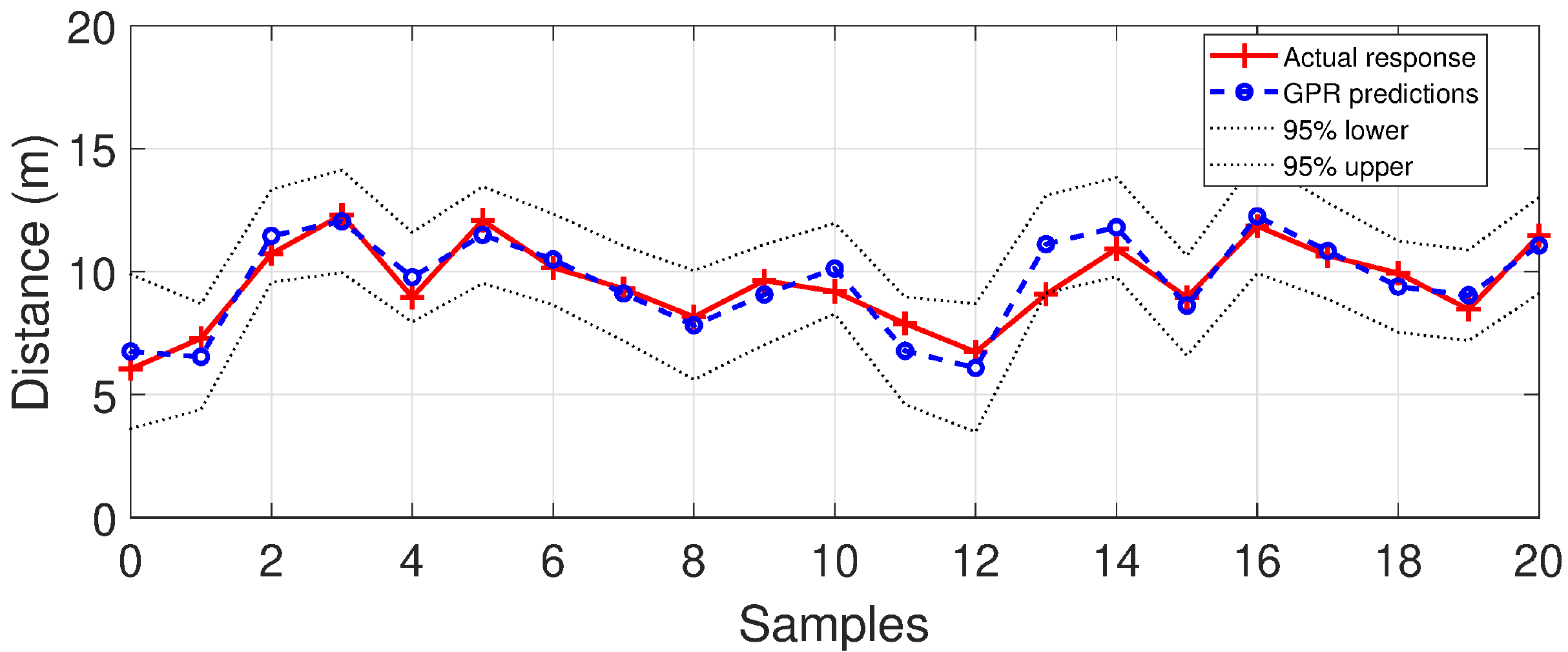} \\ 
 (b) \\
\includegraphics[width=0.95\linewidth]{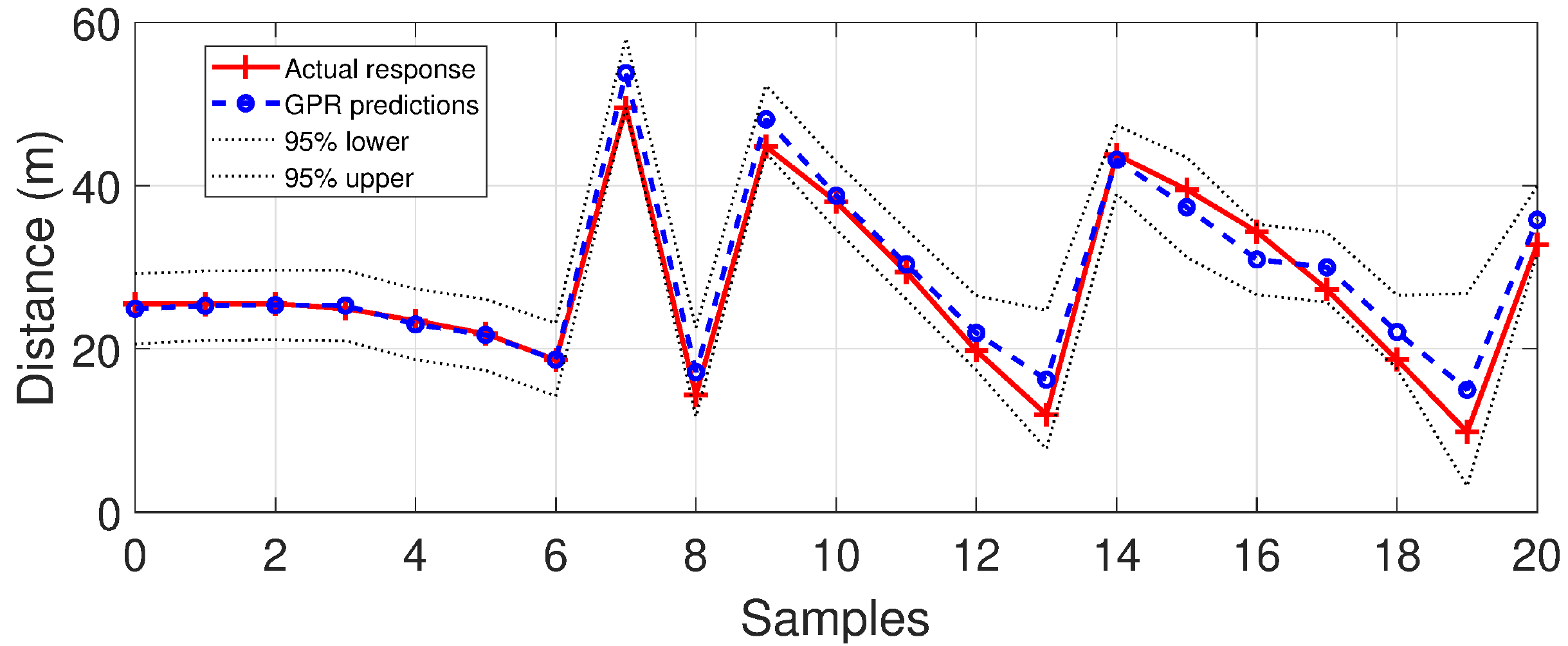} \\
\end{tabular}
\caption{Comparison between GPR performance for some samples of validation data (cross-validation): (a) Cluster of close poles and (b) Cluster of far poles.}
\label{fig:gpr_val}
\end{figure}

The GPR model trained with closer poles cluster was fed with four bounding boxes features (minimum and maximum pixels values on $x$ and $y$ image frame - $x_1, \ x_2, \ y_1, \ y_2$) and one segmentation feature (thickness of segmented pole on central part of the bounding box). The GPR model trained with far poles cluster was fed with only the four bounding boxes features. Figure \ref{fig:gpr_val} (a) and (b) presents some poles-to-vehicle distances predicted by the trained GPRs over validation data. It is interesting to observe that the target outputs are inside the GPR provided confidence bars proving that GPR is a very promising approach for this application. The mean value of the CV predicted distance absolute error, for training and validation data on both clusters were, respectively, $0.36$\,m and $0.65$\,m (close poles cluster), and $0.74$\,m and $0.98$\,m (far poles cluster).

\begin{figure}[t]
\centering
\begin{tabular}{c}
(a) \\ \includegraphics[width=0.95\linewidth]{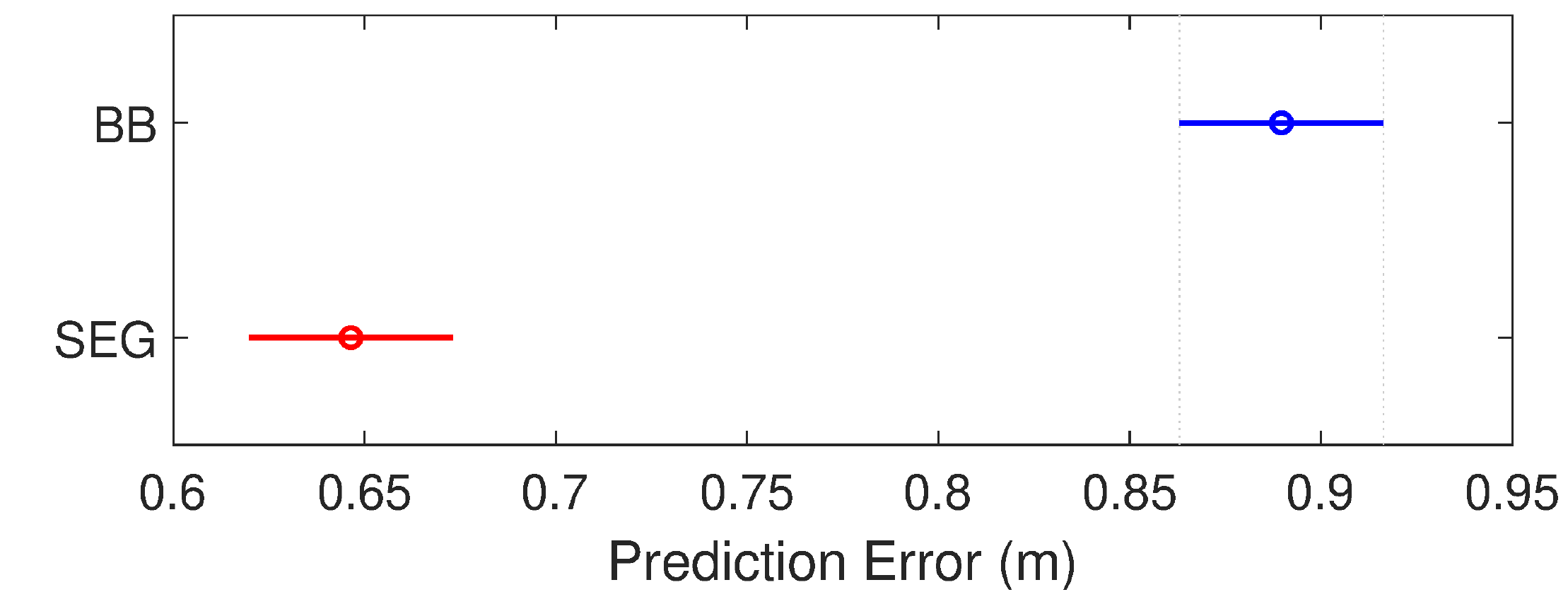} \\ 
 (b) \\
\includegraphics[width=0.93\linewidth]{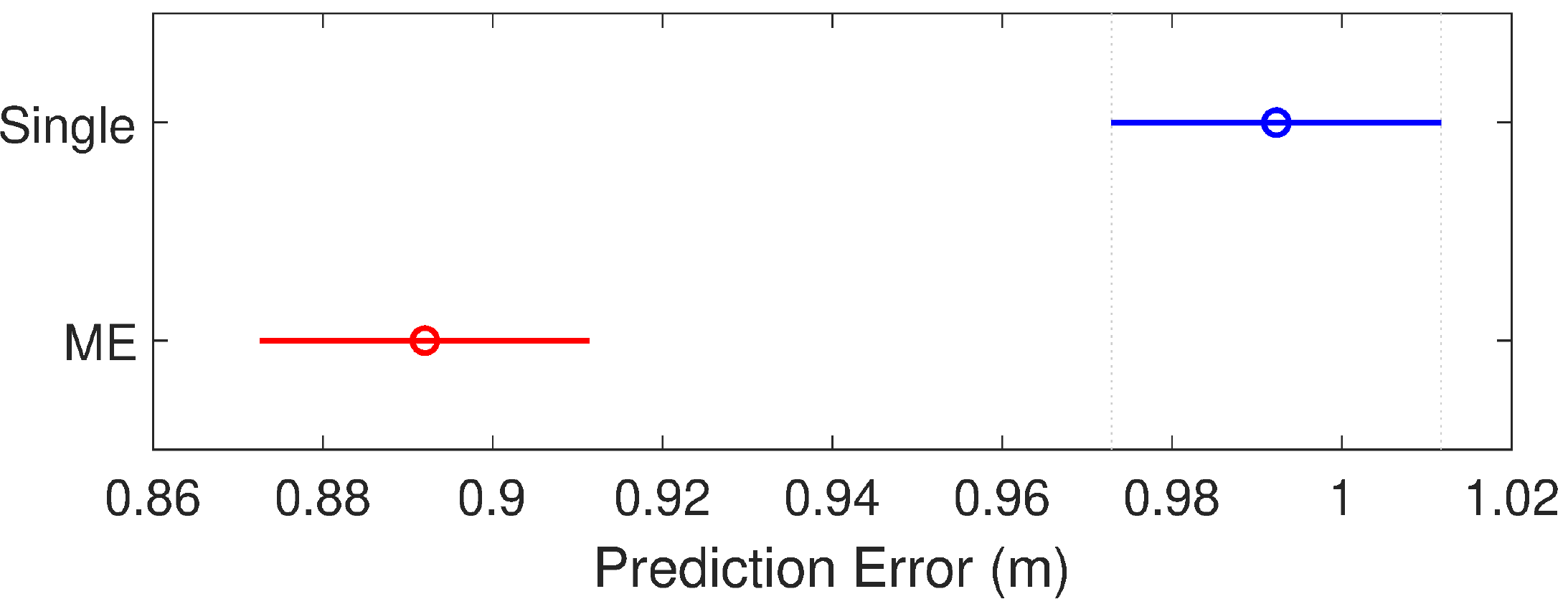} \\
\end{tabular}
\caption{Comparison of models distance prediction absolute error on cross-validation data: (a) using only bounding box  features (BB) or using bounding box together with segmentation features (SEG); and (b) using single model or mixture of expert (ME) approach.}
\label{fig:tukey}
\end{figure}

To verify that the use of segmentation feature for closer poles helped to improve the distance prediction, we compared the previous cross-validation prediction results with those from a GPR model with only Bounding Box information as inputs. A Tukey test procedure with 95\% confidence was preformed and shown in Fig.~\ref{fig:tukey} (a), from where we can observe that the extracted thickness of the poles improved the prediction accuracy. It is worth mentioning that only one simple feature based on segmentation was used in this work, we believe that adding other segmentation features may further improve the model prediction. Besides, the use of a mixture of experts (ME) is also advisable, since our ME approach outperformed a single model using the same inputs as can be inferred from Fig.~\ref{fig:tukey} (b). The use of ME is recommended for large data sets.

\subsection{Vehicle Navigation}

For the sake of simplicity, we assume that the shuttle bus dynamics can be expressed by the discrete-time \textit{nearly constant velocity} (CV) model (process model) \cite{li2003surveydyn}, 
\begin{equation}
  \mathbf{x}_{k+1} = \begin{bmatrix}
1 & 0 & T & 0\\
0 & 1 & 0 & T\\
0 & 0 & 1 & 0\\
0 & 0 & 0 & 1
\end{bmatrix} \mathbf{x}_{k} + \mathbf{w}_{k},
\end{equation}

\noindent where $\mathbf{x}_{k} = [x_k \ y_k \ \dot x_k \  \dot y_k]^\textrm{T}$ is the state vector composed of the vehicle positions and velocities in $x$ and $y$ directions, $T$ is the sampling time and $\mathbf{w}$ is a zero-mean white Gaussian noise with covariance $Q$, given by \cite{liu2019cPF}:

\begin{equation}
  Q = \begin{bmatrix}
\frac{T^3}{3} q & 0 & \frac{T^2}{2} q & 0\\
0 & \frac{T^3}{3} q & 0 & \frac{T^2}{2} q\\
\frac{T^2}{2} q & 0 & T q & 0\\
0 & \frac{T^2}{2} q & 0 & T q
\end{bmatrix},
\end{equation}

\noindent where $q$ is the process noise intensity. Besides, we consider the road surface flat, thus neglecting any height information in the vehicle movement and in the map data. It is worth mentioning that since our proposed approach is based on designing constraints for state estimation, any other vehicle motion model could be used, such as the Constant Turn Rate and Velocity (CTRV) model among many others, as described in \cite{li2003surveydyn,schubert2011models}. The same applies to the observation model here designed, where only rough GPS $x$ and $y$ position measurements are used, such that $\mathbf{z}_{k} = [x_k \ y_k] = [1\ 1\ 0\ 0]\, \mathbf{x}_{k} + \mathbf{v}_k$, and $\mathbf{v}_k$ is the sensor noise modeled as a two-dimensional zero-mean white Gaussian noise with covariance $R = \textrm{diag}(\sigma_v^2)$.

It should be noticed that the proposed approach does not depend on the implementation of any observation model, regarded the detected landmarks in the current frame are correctly matched and the vehicle initial state is known. That is, after initialization, our approach can be executed without GPS signal. Nonetheless, the observation model was implemented in our experiments to verify the influence of $R$ in the localization results and to compare the proposed approach with an unconstrained PF.    

In addition to the implementation of the aforementioned state space model, three types of soft state constraints were designed to implement the scPF: 

\begin{align}
    C_{R_k}(\mathbf{x}_k) - \gamma_{R_k} &\leq 0, \nonumber \\ 
    C_{S_k}(\mathbf{x}_k) - \gamma_{S_k} &\leq 0, \nonumber \\ 
    C_{L_{l\,k}}(\mathbf{x}_k) - \gamma_{L_{l\,k}} &\leq 0, \nonumber  
\end{align}

\noindent where $C_{R_k}(\mathbf{x}_k)$, $C_{S_k}(\mathbf{x}_k)$ and $C_{L_{l\,k}}(\mathbf{x}_k)$ are nonlinear polynomial functions of $\mathbf{x}_k$ related, respectively, to the \textit{Ring Road constraint},  the \textit{Speed constraint} and the \textit{Light poles constraints}, where $l = 1 \ldots L$ and $L$ is the number of detected landmarks. Besides, $\gamma_{R_k}, \gamma_{S_k} \textrm{and} \ \gamma_{L_{l\,k}}$ are random variables that account for the uncertainties about the road boundaries, vehicle speed limit and landmark distance prediction. It should be pointed out that many other constraints could be also included here, such as curb or lane distance prediction, some ambient signals of opportunity (SOPs) as the cellular signals, among others.

The \textit{Ring Road constraint} is used to constrain the vehicle position by the road boundaries. Considering that the centerline of the Ring Road is available, we assume that penalties would be only applied to a certain Cartesian point ($x_k$, $y_k$) if it is more than 4 meters away from the centerline. Figure \ref{fig:RingRoad_Constraints} (b) presents examples of Ring Road boundaries where the light blue area is considered unconstrained. To this end, and considering the smoothness and shape of the Ring Road, the road centerline was decomposed into a sequence of road segments represented by third order polynomials, as used in \cite{liu2019cPF}. The number of segments is the same as the number of light poles (99) since each segment represents the centerline in the vicinity of the respective light pole, such that the road constraint can be written as:

\begin{align}
    C_{R_k}(\mathbf{x}_k) & = |y_k - (b_{3}^p x_k^3 + b_{2}^p x_k^2 + b_{1}^p x_k + b^p_{0})| - 4, \nonumber \end{align}
\noindent    or
    \begin{align}
    C_{R_k}(\mathbf{x}_k) & = |x_k - (b^p_{3} y_k^3 + b^p_{2} y_k^2 + b^p_{1} y_k + b^p_{0})| - 4, \nonumber
\end{align}

\noindent where $\mathbf{b}^p = [b_{3}^p \ \, b_{2}^p \ \, b_{1}^p \ \, b_{0}^p]$ represents the coefficients of the segment $p$.  Thus, the coefficients to be used in the constraint are those related to the light pole closer to the last estimated vehicle position. The interested reader could refer to other papers in order to apply more efficient and  accurate approaches \cite{chen2010,schindler2013}. Since we are using soft constraints, the localization approach may deal with some road network uncertainties. In this regard, $\gamma_{R_k}$ is defined as a random variables that follows an exponential distribution $\gamma_{R_k} \sim \mathcal{E}(\gamma_k;\mu_R)$  as presented in Eq. \ref{eq:exp}, with $\mu_R$ equals to 0.25 as suggested in \cite{liu2019cPF}.

Considering the shuttle bus has a speed limit of 12\,\textrm{m/s}, the following nonlinear function was designed,

\begin{equation}
    C_{S_k}(\mathbf{x}_k) =  \sqrt{\dot x_k^2 \ + \ \dot y_k^2} - 12, \nonumber
\end{equation}

\noindent and $\gamma_{S_k} \sim \mathcal{E}(\gamma_k;\mu_S)$ with $\mu_S$ equals to 1 as in \cite{liu2019cPF}. 

Finally, for light poles (landmarks) constraints, the distance  between the vehicle and the landmark predicted by the GPR model (Eq. \ref{eq:gpr1}) is employed, such that

\begin{equation}
    C_{L_{l\,k}}(\mathbf{x}_k) =  |\sqrt{(x_k -x_l)^2 \ + \ (y_k - y_l)^2} - \mu_{k\,\ast}|, \nonumber
\end{equation}

\noindent where $x_l$ and $y_l$ represent the detected landmark localization and $\mu_{k\,\ast}$ is the predicted distance considering the current camera frame, as presented in Fig.\,\ref{fig:RingRoad_Constraints} (b). Besides, the predicted distance variance, $\sigma^2_{k\,\ast}$ (Eq. \ref{eq:gpr2}), is used as the variance of the zero-mean truncated Gaussian distribution $\gamma_{L_{l\,k}}$ (Eq.\,\ref{eq:gauss}). It is worth mentioning that, differently from the other implemented constraints, the landmark constraints uncertainties can change over time, according to the GPR variance predictions. This is an interesting feature of the proposed approach since unforeseen situations may occur during real application, \textit{e. g.} object partial occlusion, and the predicted variance should be higher in those scenarios indicating that the predicted distance is probably not reliable. In other words, the variance ponders the use of that specific landmark constraint during the estimation of the vehicle localization.

\begin{figure}[tb]
\centering
\includegraphics[width=0.75\linewidth]{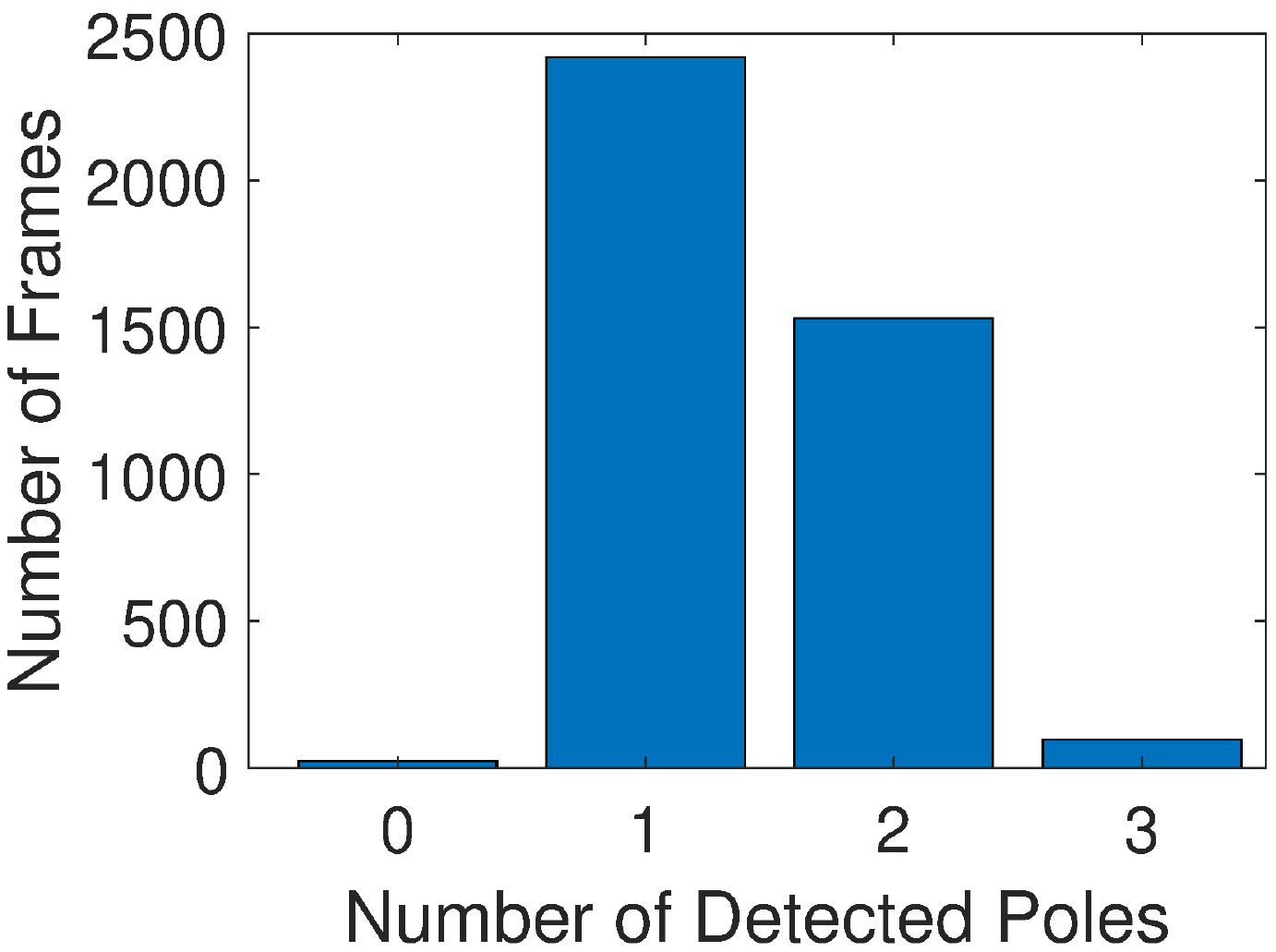}
\caption{Number of poles detected by Mask-RCNN in each image frame.}
\label{fig:hist}
\end{figure}

In order to verify the scPF performance, the acquired data set composed of 4000 samples with sampling time $T = 1/13$\,s was used.
Applying the trained Mask-RCNN model to all 4000 images, it was possible to observe that the number of poles detected varies for each frame, as shown in Fig. \ref{fig:hist}. The most frequent scenario is the detection of only one pole, although some occurrences of 2 or even 3 detected poles were found. Moreover, in some frames no poles were detected and then no landmark constraints are applied to the PF in that specific time step. 

To analyze the proposed scPF performance, 50 Monte Carlo (MC) runs were performed where different realizations of the measurement noise were generated. The proposed approach was also compared with an unconstrained PF, and both methods were tested on different measurement noise standard deviations $\sigma_v$ in order to observe the influence of this noise on algorithm's performance. To achieve the best results of each algorithm, the process noise intensities were defined as $q = 4\, \textrm{m}^2/\textrm{s}^3$ for the unconstrained PF and $ q = 11\, \textrm{m}^2/\textrm{s}^3$ for the scPF. The initial condition of the filters was achieved by the initial frame ground truth localization corrupted by noise with covariance matrix $P_0 = \textrm{diag}(10,\ 10,\ 2.5,\ 2.5)$ as in \cite{liu2019cPF}. 

\begin{figure}[tb]
\centering
\includegraphics[width=0.95\linewidth]{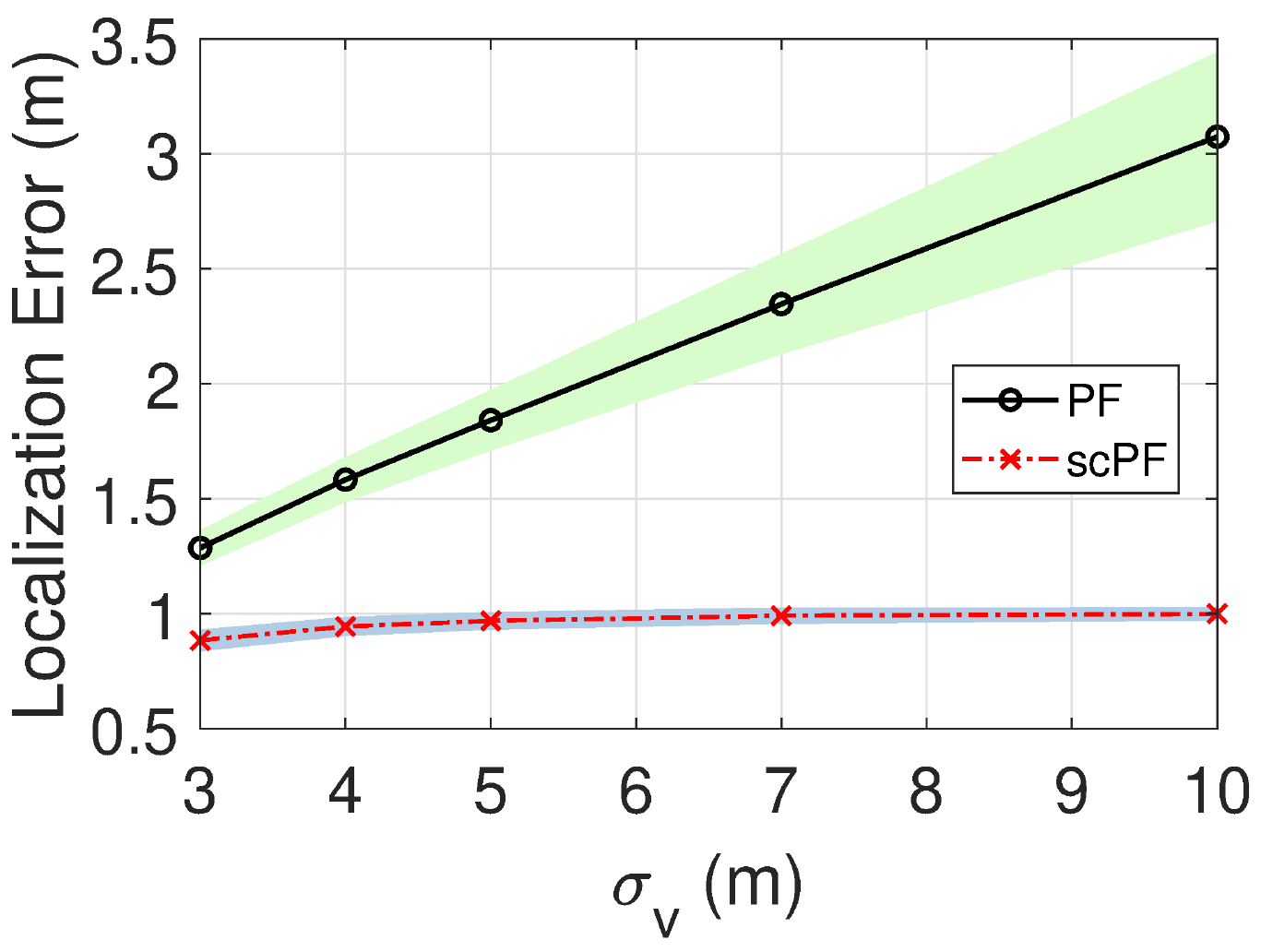}
\caption{Mean localization error and confidence levels of 95\% over 50 Monte Carlo runs of PF and scPF algorithms using different measurement noise levels.}
\label{fig:scpfpf}
\end{figure}

The results of the scPF and PF algorithms based on 500 particles over different measurement noise levels are presented in Fig. \ref{fig:scpfpf}. It is shown that the proposed scPF can reach low localization error (below 1 meter) even increasing the level of the measurement noise, which is not true for the unconstrained PF. Indeed, it can be concluded that the proposed approach does not depend on observation data since vehicle absolute localization information is available from the detected poles and the respective vehicle-to-pole distance predicted by the GPR models. Besides, when the measurement noise level is low, the localization performance is still improved, achieving a mean value of 0.88\,m when $\sigma_v = 3$\,m. 

It is worth mentioning that the proposed scPF is very promising since good localization results were found using a very simple process model and using a limited number of detected landmarks for constraints design. To make the roles of the designed constraints on the proposed approach clearer, we presented in Fig. \ref{fig:RingRoad_Constraints} (c) the sum of the constraints values (two landmarks constraints together with the road and speed constraints) for a specific time step on the testing route (Fig. \ref{fig:RingRoad_Constraints} (c) should be compared with Fig.  \ref{fig:RingRoad_Constraints} (b)). Coordinates that do not violate the constraints have dark blue colors whereas coordinates with yellow color strongly violate one or more constraints. Thus, the constraints add supplementary information to help the PF algorithm predict the vehicle localization basing on the previous system states.

\begin{figure}[tb]
\centering
\includegraphics[width=0.95\linewidth]{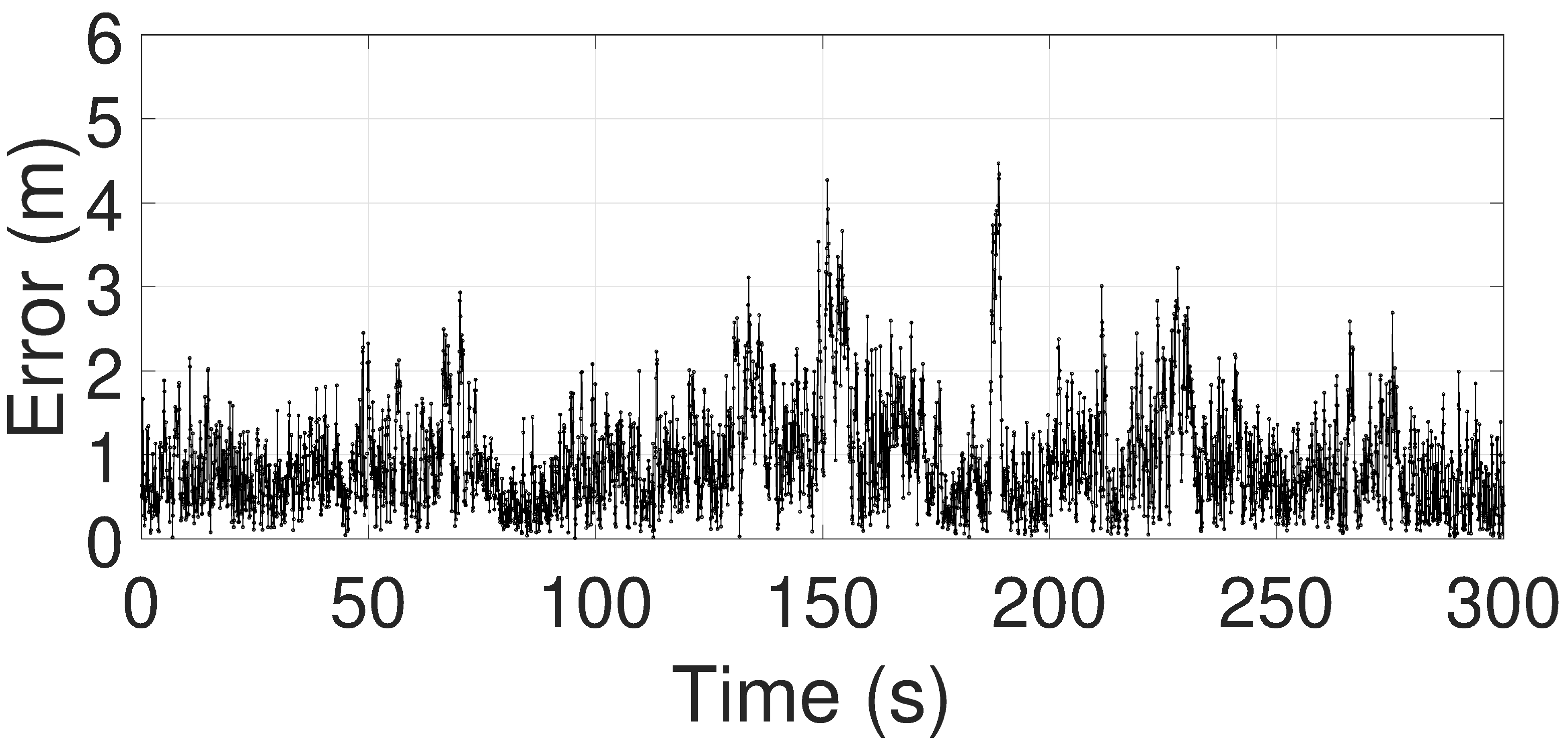}
\caption{Localization error of scPF for each time step on the Ring Road testing route.}
\label{fig:scpf300}
\end{figure}

Figure \ref{fig:scpf300} presents the localization error for each time step of the shuttle bus trajectory predicted by the scPF ($\sigma_v = 10$\, m). Some predicted localization values are also shown in Fig. \ref{fig:RingRoad_Constraints} (d). The mean error of this specific run of the scPF was 0.98 m and the median error was 0.73 m. In general, lower errors were found in frames with more landmark constraints (more poles detected), as expected. On the other hand, higher errors were achieved when no poles were detected, in these situations the localization approach relies only on the process and observation models. Since both models have high associated uncertainties due to their simplicity or measurement errors, the predicted localization error is higher. 

\begin{figure}[tb]
\centering
\includegraphics[width=0.75\linewidth]{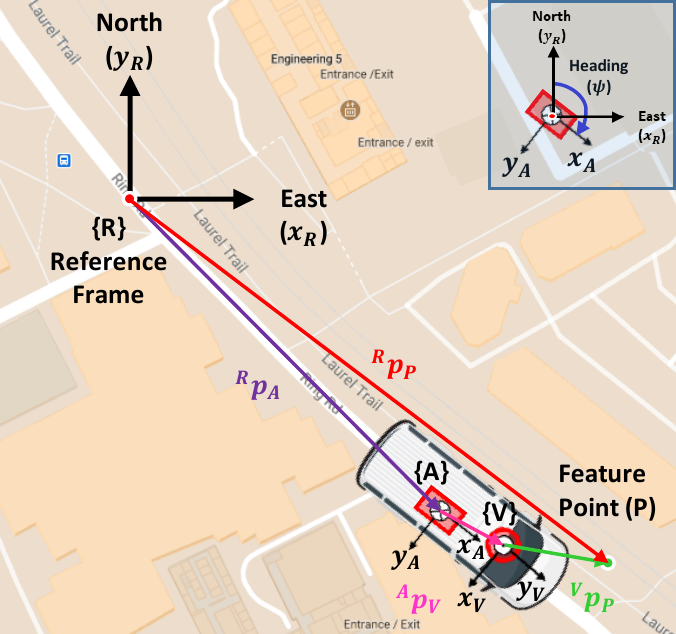}
\caption{Sensor coordinate frames as well as the position vectors including the static translation between the GPS frame \textbf{\{A\}} and the LiDAR frame \textbf{\{V\}}, and the position vector of the GPS frame \textbf{\{A\}} and the feature point of interest P in the fixed local reference frame \textbf{\{R\}}.}
\label{fig:transforms_depiction}
\end{figure}

Besides, it is worth mentioning that obtaining pole locations directly from mapping software such as Google Maps is subjected to uncertainties, which might have also worsened our algorithm performance in some frames. In order to overcome this shortcoming, 
an approach based on LiDAR measurements (if available) may be applied to obtain the position of the poles with respect to the moving LiDAR frame, \textbf{\{V\}}, and subsequently transform it to a fixed local reference frame, \textbf{\{R\}}, using a sub decimeter accurate GPS, as presented in Figure \ref{fig:transforms_depiction}. 



Thus, in order to improve the localization performance we recommend to use more mapped landmarks (for instance, traffic signs), to obtain the landmark location based on a precise method, and to implement a process model that explains the vehicle dynamics in more details.

\section{Conclusion} \label{sec:conclusion}

In this paper a generic feature-based localization approach using soft constrained Particle Filter is proposed to enhance autonomous vehicle navigation. Examples of soft constraints based on road location, on landmark-to-vehicle distance prediction and on vehicle maximum speed were provided. The use of soft constraints in this application is very promising since uncertainties are expected to be found, for instance, on road location, landmark position and landmark-to-vehicle distance prediction. Localization error below 1\,m was found even in scenarios with very noisy observation GPS data, showing that the proposed approach can be used as an add-on to other navigation approaches is such situations.

Regarding the landmark based constraints, a novel approach was introduced where Gaussian Process Regression models arranged in a Mixture of Experts scheme were obtained to predict the distance between the vehicle and known mapped landmarks. In this way, GPR predicted distance's mean and variance were both used for soft constraints designing, meaning that the constraints are continuously adapted to each frame. The output  variance provides a reliability index to the current constraint to predict the vehicle location in that specific time step, \textit{i. e.} the greater the variance the less the effect of the constraint. Moreover, in order to improve the vehicle-to-landmark distance prediction, features based on landmark instance-based segmentation were used as GPR inputs.



\bibliographystyle{elsarticle-num}
\bibliography{Refs}

\end{document}